\documentclass[runningheads]{llncs}
\usepackage[T1]{fontenc}
\usepackage{graphicx}
\usepackage{orcidlink}
\usepackage{amsmath}
\usepackage{amsfonts}
\usepackage{subcaption}
\usepackage{placeins}
\usepackage{float}
\usepackage{tikz}
\usepackage{algorithm}
\usepackage{enumitem}
\usepackage[noend]{algpseudocode} 
\usetikzlibrary{arrows.meta, positioning}

\begin{document}
\title{On-Policy Optimization of ANFIS Policies Using Proximal Policy Optimization}
\titlerunning{ANFIS Optimization through PPO}

\author{
Kaaustaaub Shankar\inst{1}\orcidlink{0009-0000-8970-3746} \and
Wilhelm Louw\inst{1}\orcidlink{0009-0009-8683-1932} \and
Kelly Cohen\inst{1}\orcidlink{0000-0002-8655-1465}
}

\institute{
College of Engineering and Applied Science, University of Cincinnati, Cincinnati, OH 45219, USA\\
\email{shankaks@mail.uc.edu,\{louwwa, cohenky\}@ucmail.uc.edu}
}

\authorrunning{K. Shankar et al.}
\maketitle              
\begin{abstract}
We present a reinforcement learning method for training neuro-fuzzy controllers using Proximal Policy Optimization (PPO). Unlike prior approaches that used Deep Q-Networks (DQN) with Adaptive Neuro-Fuzzy Inference Systems (ANFIS), our PPO-based framework leverages a stable on-policy actor-critic setup. Evaluated on the \texttt{CartPole-v1} environment across multiple seeds, PPO-trained fuzzy agents consistently achieved the maximum return of $500$ with zero variance after $20{,}000$ updates, outperforming ANFIS-DQN baselines in both stability and convergence speed. This highlights PPO's potential for training explainable neuro-fuzzy agents in reinforcement learning tasks.
\end{abstract}
\FloatBarrier
\section{Introduction}
Deep reinforcement learning (RL) has shown the potential to display super-human skill in complex domains. An example of this is AlphaGo’s defeat of a Go world champion \cite{silver2016mastering}. However, the policies learned by deep neural networks (DNNs) remain largely opaque, limiting trust in safety-critical settings such as autonomous driving and healthcare.
In contrast, fuzzy inference systems offer transparency while providing a robust solution. These systems fall into two families: Mamdani and Takagi-Sugeno-Kang (TSK).  
Mamdani systems rely on linguistic IF–THEN rules with fuzzy outputs and subsequent defuzzification, making them highly interpretable but less amenable to gradient-based tuning \cite{mamdani1975experiment}.  
TSK models instead express rule consequents as linear functions of the inputs, producing smoother numeric outputs and enabling more robust numerical optimization \cite{6313399}.
Yet both architectures still lack systematic training pipelines.  Designing membership functions, rule bases, and consequents typically depends on expert heuristics or search methods such as genetic algorithms, which hampers scalability to high-dimensional or dynamic tasks \cite{CORDON}.

Neuro-fuzzy methods like ANFIS address this by using a neural network to transform the inputs into intermediate features, which feed into Gaussian membership functions that activate first-order TSK rules; their weighted outputs are aggregated to produce the final action logits. All trainable parameters like network weights, membership centres and sigmas, and rule consequents are updated through gradient descent \cite{jang1993}. Furthermore, deep applications like ANFIS-DQN hybrids have shown promise \cite{zander2023reinforcement} but inherit the instability of off-policy Q-learning.

Proximal Policy Optimization (PPO) combats these issues with a clipped, on-policy surrogate objective that yields stable learning and strong sample efficiency \cite{schulman2017proximalpolicyoptimizationalgorithms}. We therefore integrate an ANFIS-style fuzzy module into PPO, forming a \emph{PPO-Fuzzy} agent. Using the well-studied \texttt{CartPole-v1} benchmark, we evaluate whether the approach achieves the transparency of fuzzy rules without sacrificing the performance of modern policy-gradient RL.

\FloatBarrier
\section{Related Work}
Recently, there has been growing interest in scaling up fuzzy RL and integrating it with deep learning. Zander \emph{et al.}\ trained Takagi–Sugeno–Kang (TSK) fuzzy systems with Deep Q-Learning (DQN), reporting \texttt{CartPole-v1} performance on par with, or better than, ordinary DQNs yet exhibiting the training instabilities typical of off-policy methods \cite{zander2023reinforcement}. This motivates exploring \emph{on-policy} optimization such as PPO.
\FloatBarrier
\section{Methodologies}
To isolate the effect of the optimization algorithm, we replicate the experimental setup of Zander \emph{et al.}, replacing their DQN learner with a PPO-based actor–critic loop. Four agents, each initialized with a distinct random seed, are trained on the \texttt{CartPole-v1} environment.

As in the original work, the raw state vector is passed through a neural network comprising an input layer with 4 units, a hidden layer of 128 neurons with \texttt{ReLU} activation, and a second hidden layer of 127 \texttt{ReLU}-activated neurons. These 127 intermediate features are used to activate 16 Gaussian membership functions, whose firing strengths are used in a weight sum of the first-order TSK rule consequents to produce the final action.

\begin{figure}[H]
  \centering
  \includegraphics[width=0.8\linewidth]{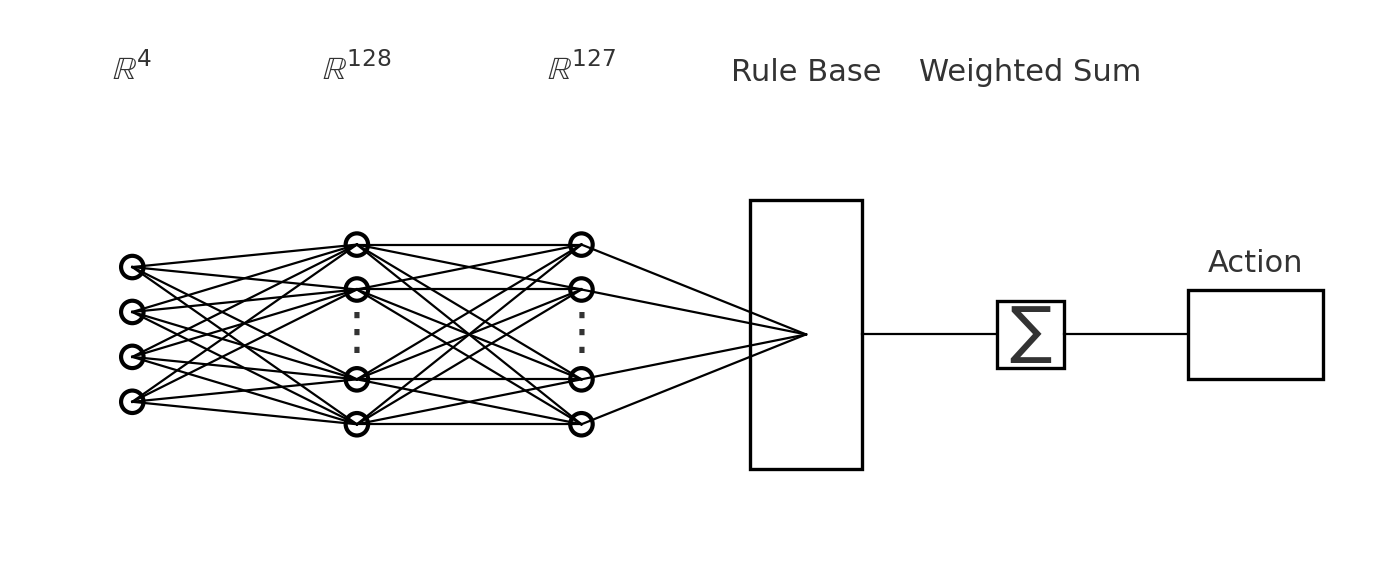}
  \caption{Architecture of the controller}
  \label{fig:controller}
\end{figure}

The value function, used to estimate state values for PPO’s critic, is modeled by a separate neural network. This network consists of two fully connected hidden layers with 64 and 32 units respectively, each followed by a \texttt{Tanh} activation.

\begin{figure}[H]
  \centering
  \includegraphics[width=0.7\linewidth]{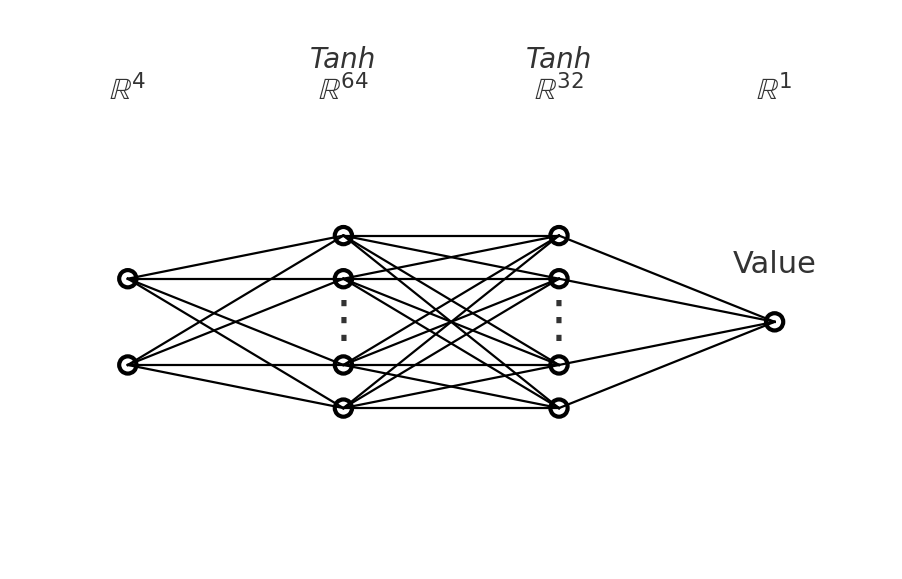}
  \caption{Architecture of the value function}
  \label{fig:value}
\end{figure}

The actor and critic networks are optimized jointly using the PPO objective, which balances clipped policy updates, value regression, and entropy regularization to ensure stable learning. The full training loop for the PPO-ANFIS agent is summarized in Algorithm~\ref{alg:ppo_anfis}.

\begin{algorithm}[!htbp]
\caption{PPO-ANFIS training loop}\label{alg:ppo_anfis}
\begin{algorithmic}[1]
\State Initialise ANFIS policy $\pi_{\theta}$ and value network $V_{\phi}$
\For{each iteration $i=1,\dots,N_{\text{updates}}$}
    \State Collect $T$ timesteps of on-policy data $\mathcal{D}_i$ using $\pi_{\theta}$
    \State Compute returns $R_t$ and advantages $\hat A_t$ for all $(s_t,a_t)\in\mathcal{D}_i$
    \For{epoch $k=1$ \textbf{to} $K$}
        \State Sample minibatch $\mathcal{B}\subset\mathcal{D}_i$
        \State $\mathcal{L}_{\text{clip}} = 
        \mathbb{E}_{\mathcal{B}}\!\Big[
           \min\!\big(r_t(\theta)\hat A_t,\;
           \operatorname{clip}(r_t(\theta),1-\epsilon,1+\epsilon)\hat A_t\big)
        \Big]$
        \State $\mathcal{L}_{\text{VF}} = \frac12(R_t-V_{\phi}(s_t))^{2}$\hfill(Value loss)
        \State $\mathcal{L} = -\mathcal{L}_{\text{clip}} + c_{v}\mathcal{L}_{\text{VF}} - c_{e}\,\mathbb{E}_{\mathcal{B}}[H[\pi_{\theta}(\cdot|s_t)]]$
        \State Update $(\theta,\phi)$ via Adam; clip gradient-norm to $10$
    \EndFor
\EndFor
\end{algorithmic}
{\textbf{Notation.} $r_t(\theta)=\pi_{\theta}(a_t|s_t)/\pi_{\theta_{\text{old}}}(a_t|s_t)$; $c_{v}$ and $c_{e}$ are value and entropy weights; $\epsilon=0.2$ is the PPO clip parameter.}
\end{algorithm}

All experiments use the \texttt{CartPole-v1} environment with 
the episode length capped at 500 steps. We use four runs that employ the following seeds $\{9,\allowbreak 42,\allowbreak 109,\allowbreak 131\}$; each run trains for $1{\times}10^{5}$ mini-batch updates. The hyperparameters are included in the Appendix.

\FloatBarrier
\section{Results}           
\begin{figure}[H]          
  \centering

  \begin{subfigure}{0.8\linewidth}
    \centering
    \includegraphics[width=\linewidth]{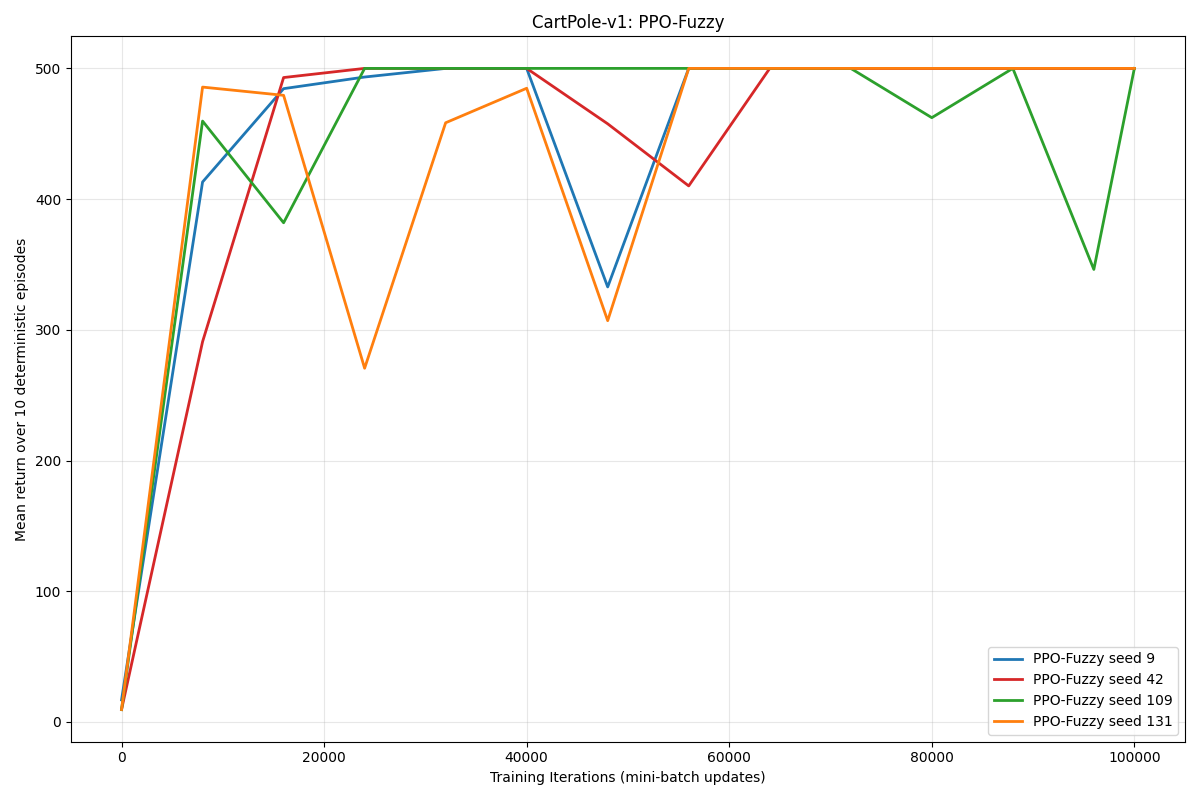}
    \caption{PPO-trained ANFIS agents on \texttt{CartPole-v1}. 
             Mean return over 10 deterministic episodes, averaged across seeds.}
    \label{fig:ppo_eval1}
  \end{subfigure}

  \begin{subfigure}{0.8\linewidth}
    \centering
    \includegraphics[width=\linewidth]{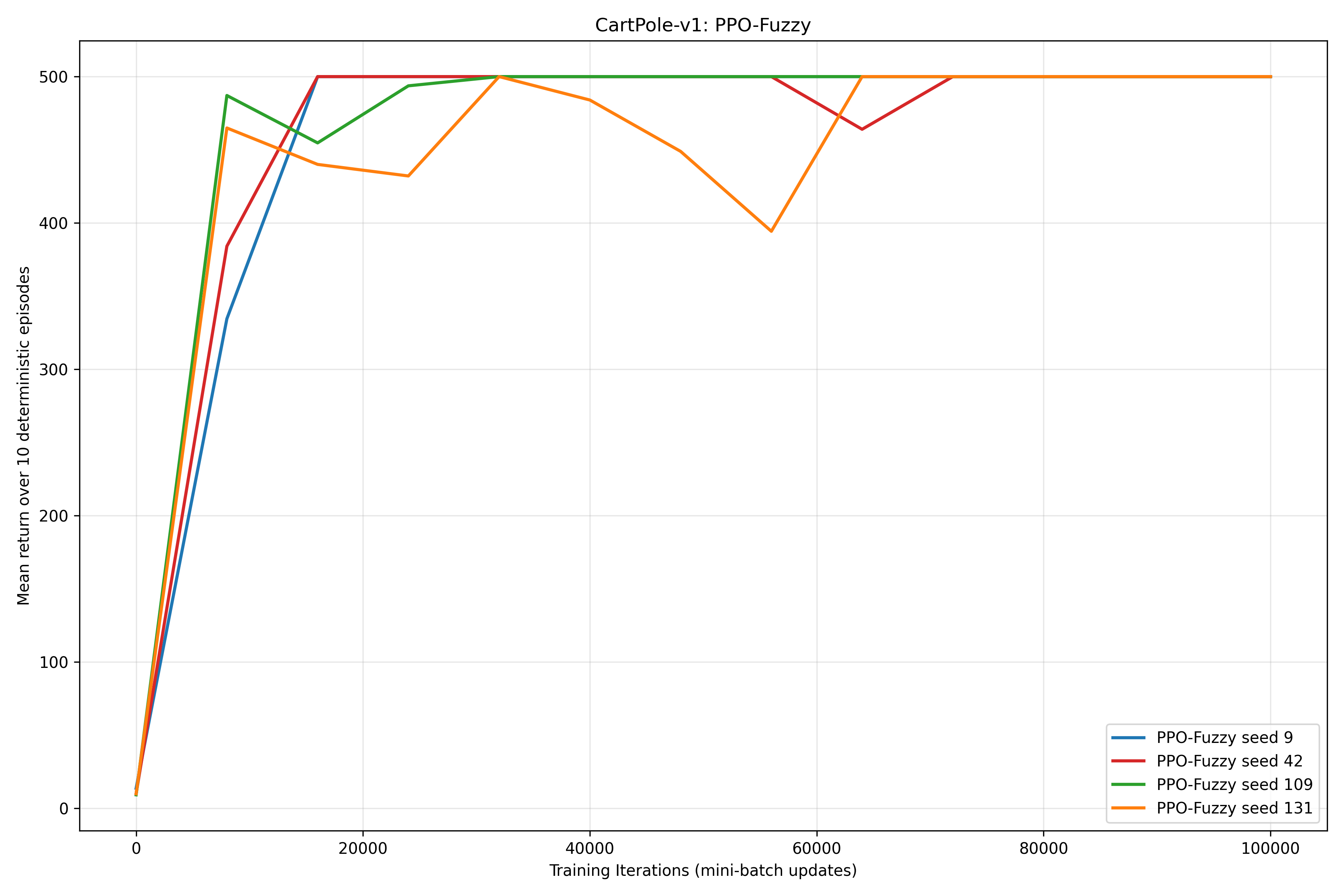}
    \caption{Same experiment with gradient-norm clipping at 0.5.}
    \label{fig:ppo_eval2}
  \end{subfigure}

  \caption{Evaluation curves for PPO-trained ANFIS controllers.}
  \label{fig:ppo_eval_both}
\end{figure}

\begin{figure}[H]
  \centering
  \includegraphics[width=0.6\linewidth]{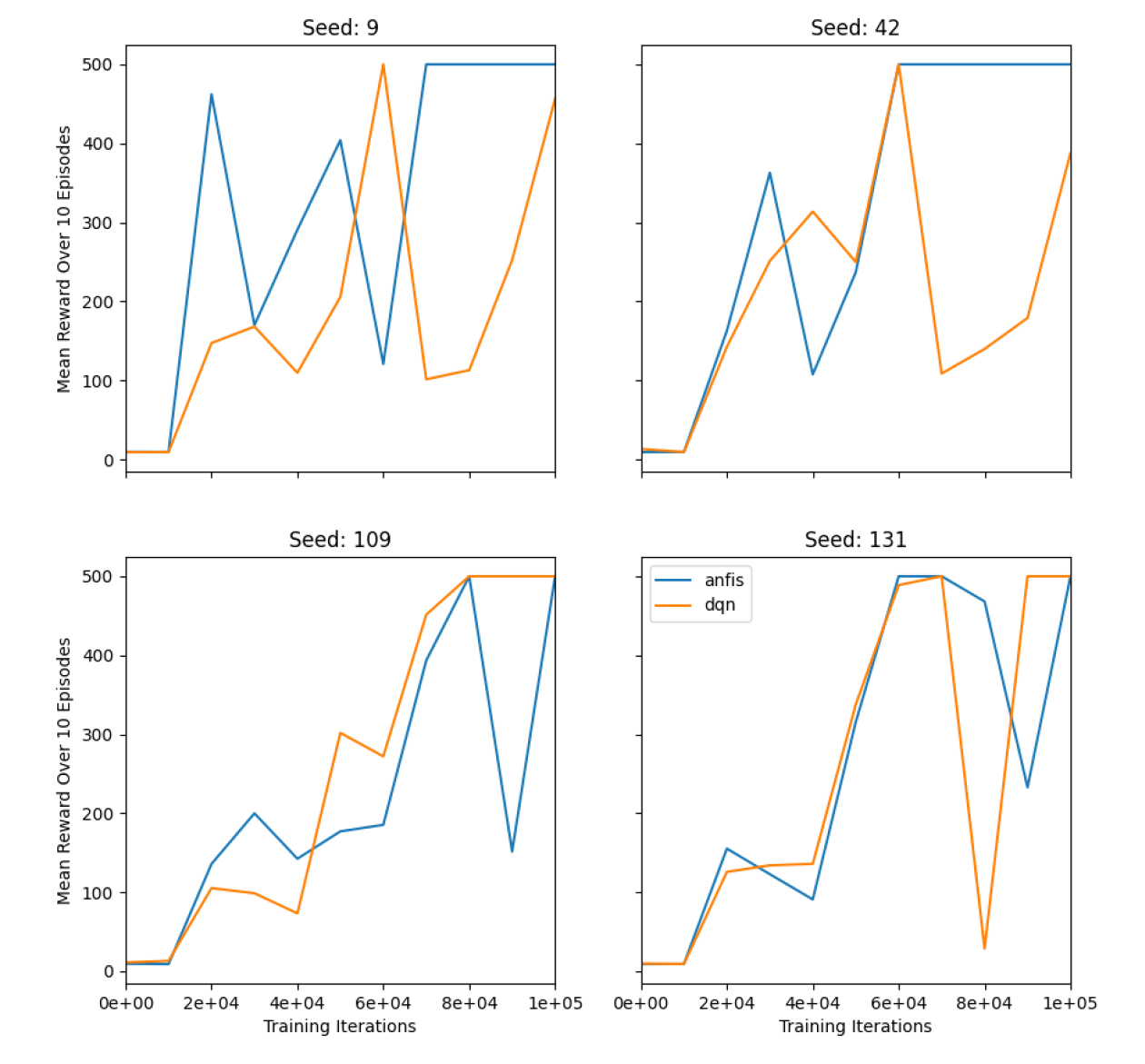}
  \caption{ANFIS-DQN and vanilla DQN agents on \texttt{CartPole-v1} 
           (Referenced from \cite{zander2023reinforcement}).}
  \label{fig:dqn_eval}
\end{figure}
\FloatBarrier
\section{Discussion}
Figure~\ref{fig:ppo_eval1} shows the training performance of PPO-trained ANFIS agents in the \texttt{CartPole-v1} environment across four random seeds, while Figure~\ref{fig:dqn_eval} presents the results for ANFIS-DQN and standard DQN. All PPO-based agents converge rapidly, reaching the maximum return of 500 within 20,000–40,000 mini-batch iterations.

Some fluctuations remain (seeds 9 and 131 around 50,000 iterations), partly due to hyperparameter choices. With more optimal settings such as gradient-norm clipping at 0.5, training stability improves as shown in Figure~\ref{fig:ppo_eval2}. By 100,000 iterations, all PPO agents consistently achieve the maximum return, demonstrating robustness to initialization and effective handling of the fuzzy policy structure.

In contrast, prior DQN-based ANFIS training~\cite{zander2023reinforcement} showed persistent instability. These results highlight PPO's suitability for training fuzzy controllers and its potential for scalable, stable learning in higher-dimensional tasks.
\FloatBarrier
\section{Future Work}
In future work, we aim to expand and test this framework to more complicated environments like \texttt{LunarLander-v3-Continuous} and \texttt{Hopper-v1}. We also plan to explore integration of interpretability tools like SHAP or LIME to attribute actions to specific fuzzy rules, guiding rule pruning and the discovery of the optimal rule count.
\FloatBarrier
\begin{credits}
\subsubsection{\ackname}
The authors extend their sincere gratitude to the members of the AI Bio Lab at the University of Cincinnati for their invaluable discussions and collaborative efforts that facilitated the realization of this work. The authors especially thank Bharadwaj `Ben' Dogga for his thoughtful review of early drafts.

\subsubsection{\discintname}
The authors have no competing interests to disclose.
\end{credits}
\FloatBarrier
\bibliographystyle{unsrt}
\bibliography{Bib}
\FloatBarrier
\section{Appendix:}
\begin{table}[H]
  \centering
  \label{tab:hyperparams_simple}

  \begin{tabular}{|l|l|}
    \hline
    \textbf{Parameter} & \textbf{Value} \\
    \hline
    Discount factor $\gamma$         & $0.99$ \\
    Learning rate                    & $1\times10^{-5}$ \\
    PPO clip $\epsilon$              & $0.2$ \\
    Entropy coefficient $c_{e}$      & $0.02$ \\
    Mini-batch size                  & $64$ \\
    Roll-out horizon $T$             & 2048 steps \\
    Gradient-norm clip               & 10 \\
    Membership centres $c_i$         & $\mathcal N(0,0.1^{2})$ \\
    Membership widths $\sigma_i$     & $0.25+0.5\,\mathcal U(0,1)$ \\
    Rule consequents $w_i$           & $2\,\mathcal N(0,1)$ \\
    \hline
  \end{tabular}
\end{table}

\end{document}